\documentclass{article}

\PassOptionsToPackage{numbers, compress}{natbib}

\usepackage[preprint]{neurips_2024}

\usepackage[utf8]{inputenc} 
\usepackage[T1]{fontenc}    
\usepackage{url}            
\usepackage{booktabs}       
\usepackage{amsfonts}       
\usepackage{nicefrac}       
\usepackage{microtype}      
\usepackage{xcolor}         
\usepackage{hyperref}

\usepackage{graphicx}
\usepackage{subfigure}
\usepackage{amsmath}
\usepackage{mathtools}
\usepackage{amsthm}
\usepackage{fancyhdr}       
\usepackage{graphicx}       
\usepackage{listings}
\usepackage{multirow}
\usepackage{multicol}
\usepackage{bm}
\usepackage{amsmath}
\usepackage{xspace}
\usepackage{algorithm}
\usepackage{wrapfig}
\usepackage[capitalise,noabbrev]{cleveref}

\makeatletter
\AfterEndEnvironment{algorithm}{\let\@algcomment\relax}
\AtEndEnvironment{algorithm}{\kern2pt\hrule\relax\vskip3pt\@algcomment}
\let\@algcomment\relax
\newcommand\algcomment[1]{\def\@algcomment{\footnotesize#1}}
\makeatother
\usepackage{algpseudocode}
\theoremstyle{plain}

\theoremstyle{definition}

\theoremstyle{remark}

\usepackage[textsize=tiny]{todonotes}
\usepackage{xcolor}

\newcommand{\ie}{\textit{i.e.},\xspace}

\newcommand{\eg}{\textit{e.g.},\xspace}
\newcommand{\bx}{\bm{x}}

\newcommand{\bz}{\bm{z}}

\newcommand{\bs}{\bm{s}}

\newcommand{\btheta}{\bm{\theta}}

\newcommand{\ms}{\mathcal{S}}

\newcommand{\emb}{text embeddings}

\title{Federated Generative Learning with Foundation Models}

\author{Jie Zhang \\
ETH Zurich \\
\texttt{jie.zhang@inf.ethz.ch} \\
\And
Xiaohua Qi \\
USTC \\
\texttt{xhqi@mail.ustc.edu.cn} \\
 \And
Bo Zhao \\
BAAI \\
\texttt{zhaobo@baai.ac.cn} \\
}

\begin{document}

\maketitle

\begin{abstract}
Existing approaches in Federated Learning (FL) mainly focus on sending model parameters or gradients from clients to a server. 
However, these methods are plagued by significant inefficiency, privacy, and security concerns. Thanks to the emerging foundation generative models, we propose a novel federated learning framework, namely \emph{Federated Generative Learning}. 
In this framework, each client can create \emb{} that are tailored to their local data, and send embeddings to the server. Then the informative training data can be synthesized remotely on the server using foundation generative models with these embeddings, which can benefit FL tasks.
Our proposed framework offers several advantages, including \textbf{increased communication efficiency}, \textbf{robustness to data heterogeneity}, \textbf{substantial performance improvements}, and \textbf{enhanced privacy protection}. We validate these benefits through extensive experiments conducted on 12 datasets. For example, on the ImageNet100 dataset with a highly skewed data distribution, our method outperforms FedAvg by 12\% in a single communication round, compared to FedAvg's performance over 200 communication rounds. We have released the code for all experiments conducted in this study\footnote{\href{https://github.com/zj-jayzhang/Federated\_Generative\_Learning}{https://github.com/zj-jayzhang/Federated\_Generative\_Learning} }.
\end{abstract}

\section{Introduction}
Recently, significant progress has been achieved in many learning fields by scaling up to large models, \ie BERT~\cite{devlin2018bert}, GPT3~\cite{brown2020language}, ViT~\cite{dosovitskiy2020image}, CLIP~\cite{radford2021learning}, Stable Diffusion~\cite{Rombach_2022_CVPR}, and Web-scale datasets \ie YFCC100M~\cite{YFCC100M}, CC-12M~\cite{changpinyo2021cc12m}, LAION-5B~\cite{LAION5B}. Typically, large models are first pre-trained with massive low-quality web data for basic capability, then finetuned with a small number of high-quality data, especially manually labeled data, for evoking the desired capability. 
Although Web data are easily accessible, high-quality training data remains scarce due to the fact that high-quality datasets are typically private or unsuitable for public release.
For example, the process of labeling medical data is often costly, and the release of such data is sensitive due to safety and privacy concerns. Furthermore, raw data itself are often considered a valuable asset for numerous companies, rendering its acquisition impractical. 
Consequently, there is a pressing need for collaborative machine learning~\cite{gong2022preserving,nguyen2022preserving,mothukuri2021survey} that is efficient and privacy-preserving.
\begin{figure*}[t]
    \centering
    \vspace{-2mm}
    \subfigure[Test accuracy on 8 datasets in one round.]{
        \includegraphics[width=0.48\textwidth]{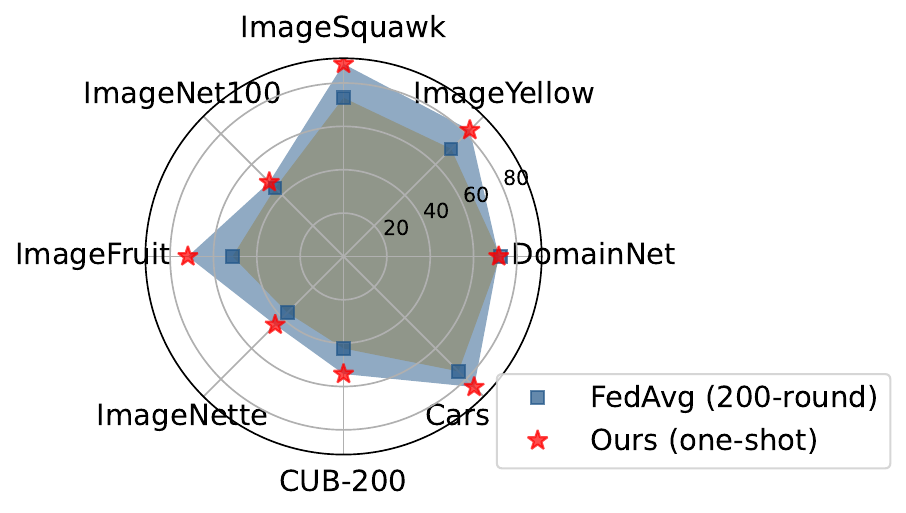}
        \label{fig:intro_a}
    }
    \subfigure[Results on satellite datasets and medical datasets.]{
        \includegraphics[width=0.45\textwidth]{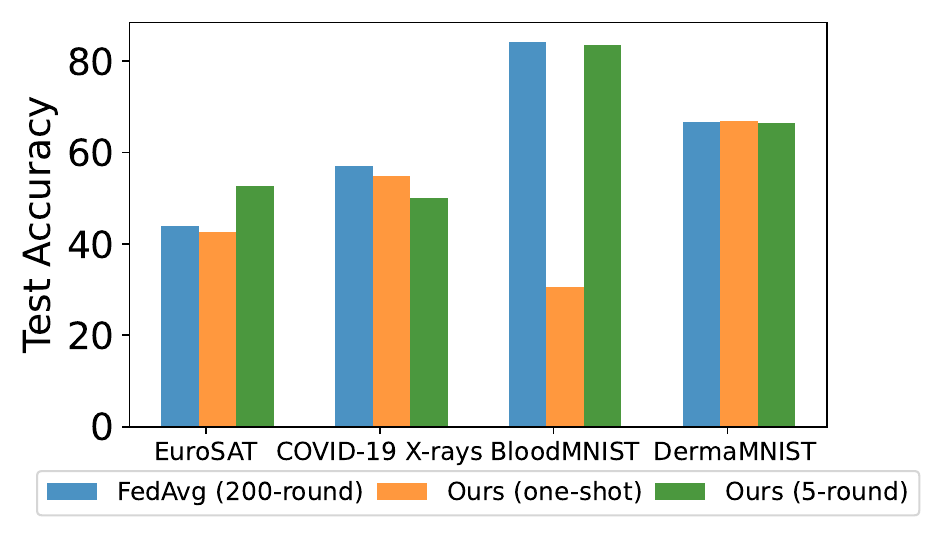}
        \label{fig:intro_b}
    }
     \caption{For datasets such as subsets of ImageNet or DomainNet, our proposed method can achieve superior accuracy with only a single round of communication. In scenarios involving inherently challenging domains, including medical datasets and satellite imagery, our approach can still attain comparable performance with only five rounds of communication. }
    \label{fig:intro}
    \vspace{-4mm}
\end{figure*}

Federated Learning (FL)~\cite{fedavg} has been gaining a lot of attention as a potential way to protect user privacy in distributed machine learning. When it comes to practical applications, FL systems face a few challenges that impede their real-world implementation:
\begin{enumerate}
    \item  \emph{\textbf{High communication cost}}. Current FL solutions require the transmission of model parameters or gradients between clients and the server~\cite{fedavg,zhao2018federated}. However, in the era of large models, these parameters are often in the billions or trillions, making their communication costly and even prohibitively expensive.
    \item \emph{\textbf{Data heterogeneity}}. In FL, a fundamental challenge arises from the presence of statistical heterogeneity among local data distributions across distinct clients, which leads to  performance degradation~\cite{wang2020tackling,zhang2022federated}.
    \item \emph{\textbf{Privacy and security risks}}. Traditional FL involves the transmission of model parameters or gradients between clients and server. However, once these model parameters are leaked, attackers can carry out model extraction attacks~\cite{li2023model}, member inference attacks~\cite{li2023effective}, and other malicious activities~\cite{zhu2019deep,advfl}, posing significant security threats to FL system~\cite{nguyen2022preserving,liu2022threats}.  
\end{enumerate}

Recent advances in foundation generative models, \ie Stable diffusion ~\cite{rombach2022high}, DALL-E2 ~\cite{ramesh2022hierarchical}, Imagen ~\cite{saharia2022photorealistic}, and GLIDE ~\cite{nichol2021glide}, have provided a high-quality conditional text image synthesis that can be used to train models.  These foundation generative models have been applied in various fields, including Computer Vision~\cite{blattmann2023align,ruiz2023dreambooth},  Speech~\cite{liu2023diffvoice,levkovitch2022zero}, and have achieved remarkable results.



In this paper, we propose a novel framework called \textbf{\textit{Federated Generative Learning} (FGL)}, which leverages powerful foundation generative models, \eg Stable Diffusion~\cite{Rombach_2022_CVPR}, to synthesize high-quality training data on the server based on the \emb{} collected from clients. We propose two customized prompt generation methods based on the characteristics of the client's data, and these prompts are then used as inputs to a specific text encoder to obtain corresponding \emb{}.
Once all \emb{} are collected from the clients, the server performs embedding aggregation and then synthesizes a high-quality substitute training dataset. This public synthetic dataset serves as a proxy for the clients' private data and can be used to train a global model on the server.  In~\cref{fig:intro_a}, our trained model in a single round outperforms FedAvg with 200 communication rounds on 8 popular datasets. We further demonstrate the effectiveness of FGL by presenting results on more complex satellite dataset and three medical datasets in~\cref{fig:intro_b}.
In summary, there are multiple benefits of FGL: 
\begin{enumerate}
    \item \emph{\textbf{Low Communication Cost}}. Compared to previous methods that rely on multi-round communication of model parameters or gradients, our method requires only one or a few communication rounds (\eg 5 rounds) between clients and the server. Despite this efficiency, our method is able to achieve performance that is on par with the existing methods. 
    \item \emph{\textbf{Robust to Data Heterogeneity}}. Since our method only requires clients\footnote{FGL requires all clients to participate in the initial round to get all text embedding first on the server. It is more suitable for cross-silo FL~\cite{huang2022cross} scenarios, where the clients represent organizations or companies, and the number of clients is typically small.} to upload \emb{} corresponding to their local training data, it allows the server to collect embeddings from all clients and synthesize all training data in one communication round. Based on the well-synthesised data, our method exhibits insensitivity to the data distribution. 
    
    \item \textit{\textbf{Better privacy-preserving}}: Previous FL methods are vulnerable to various attacks because they always transmit model parameters/gradients during the learning process. In contrast, our method only transmits \emb{} in the first round\footnote{In subsequent rounds, FGL also transmits model parameters or gradients. However, the model memorizes very little private information after only a few rounds, which mitigates privacy risks compared to conventional FL methods based on multi-round communications. See detailed experiments on privacy analysis in Section~\ref{sec:privacy}.}. In ~\cref{sec:privacy}, we conduct a thorough privacy analysis on two aspects: (1) whether the synthetic data reveals private data information visually and (2) whether the model trained on the synthetic data is resilient against privacy attacks (e.g., membership inference attack~\cite{shokri2017membership}).

\end{enumerate}
\begin{figure*}[t]  
    \centering
    \includegraphics[width=14cm]{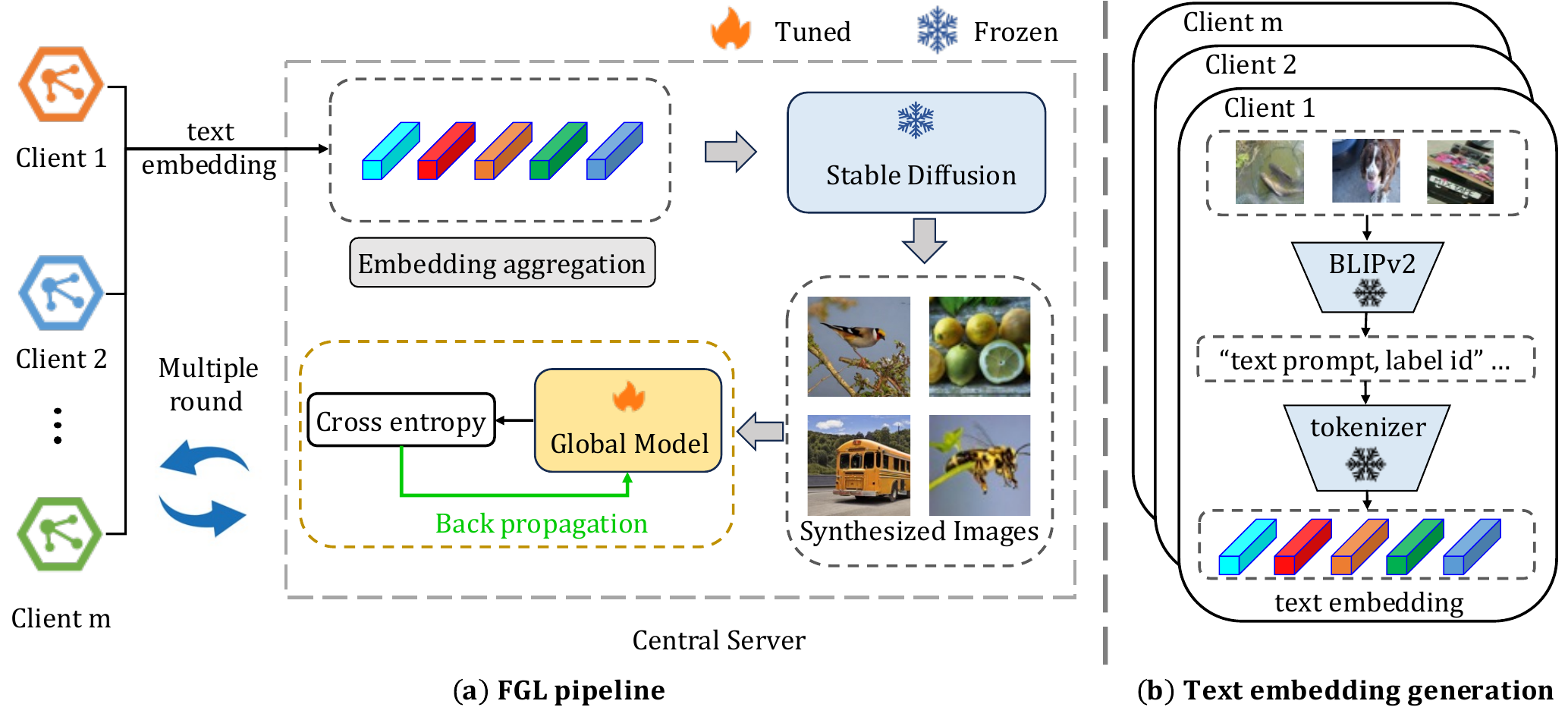}
    \vspace{-5mm}
   \caption{Training pipeline of FGL. Firstly, the \emb{} from clients are uploaded and then aggregated on the server. Then, stable diffusion is used to generate synthetic data to train the global model. Finally, the updated model are distributed to all clients. In the subfigure on the right, we present a detailed process of generating text embeddings.}
 
    \label{fig:pipeline}
    \vspace{-4mm}
\end{figure*}
\section{Federated Generative Learning}
\label{sec:fgl}
\begin{wrapfigure}{r}{0.46\textwidth}
\vspace{-8mm}
\begin{minipage}{0.47\textwidth}
\begin{algorithm}[H]
\caption{\small Federated Generative Learning}
\label{alg:code}
\algcomment{
\textbf{Notes}: 
We implemented FGL in both a single communication round and five communication rounds.
}
\definecolor{codeblue}{rgb}{0.25,0.5,0.5}
\definecolor{codekw}{rgb}{0.85, 0.18, 0.50}
\begin{lstlisting}[language=python, basicstyle=\fontsize{8pt}{9.6pt}\selectfont]
# 1) first round, create prompts and embeddings.
embed_list = []
for client in all_clients:
    # given data (x,y), generate prompts
    # class-level or instance-level
    prompts, y_s = prompts_generation(client)
    embed_list.append(text_encoder(prompt),y_s)
# 2) generate data and train global model
embeds, y_s = embeds_aggregation(embed_list)
syn_data = generative_model(embeds)
global_model = server_update(syn_data, y_s) 
# 3) for 5-round communication
for com_round in [2,3,4,5]:
    local_weights = []
    for client in selected_clients:
        weight = local_update(client, global_model) 
        local_weights.append(weight)
    global_model = model_averaging(local_weights)
    # finetuning for highly skewed data on server.
    if server_finetune: server_update(syn_data, global_model) 
\end{lstlisting}
\end{algorithm}
\end{minipage}
\vspace{-6mm}
\end{wrapfigure}
\paragraph{Framework Overview} The overall framework of the proposed \emph{Federated Generative Learning} framework is illustrated in Figure ~\ref{fig:pipeline}. Unlike traditional FL methods that transmit features, parameters, or gradients, our approach transmits \emb{} corresponding to the private data in clients to the server, thus being better privacy-preserving and communication-efficient. Then the training data is synthesized based on the aggregated \emb{} in the server with the foundation diffusion model. The synthetic training data are jointly used to train models, thus can relieve data heterogeneity problem and improve performance. Our framework has the capability to execute one-shot FL, eliminating the need for clients to train models on their local devices. Additionally, we demonstrate an extension that incorporates five communication rounds, resulting in better results. Since the generative model has never seen any private data from clients, and all synthetic data is generated and stored on the server, FGL does not necessitate additional computation resources on the client-side and does not compromise clients' privacy.
The Pytorch-like pseudocode of our method is presented in~\cref{alg:code}.

\subsection{Federated Learning Setting} 
In FL, we have $K$ clients with their private datasets $\mathcal{D}_k = \{(\bx_i, y_i)\}_{i=1}^{N_k}$, where $\bx_i$ is the training image, $y_i$ is its label, $\mathcal{Y}_k$ is the label set, and $N_k$ is the number of training samples in $k$-th client. Note that the label sets of different clients may be different. The objective of the federated learning framework is to learn a model parameterized with $\btheta$ in the server that minimizes the loss on training data of all clients without access to original data: $\min_{\btheta} \frac{1}{K} \sum_{k=1}^{K} \mathbb{E}_{\bx \sim \mathcal{D}_k} [l_k(\btheta; \bx)],$
where, $l_k$ is the loss function for the $k$-th client.
In this study, we explore the scenario wherein each client transfers text embeddings to the server. Subsequently, we leverage a foundational generative model on the server to generate a set of synthetic data, which can be utilized for pretraining a global model.

\subsection{Text Embedding Generation and Aggregation}
\label{sec:prompt aggregation}
We investigate two types of prompt generation: class-level prompt and instance-level prompt:
\begin{itemize}
    \item The class-level prompts are generated based on class names, providing high-level guidance to the generative model. For example, for each client, we generate a prompt like `A photo of a \{class name\}' for each data. Various images with different types of noise can be generated using each prompt at the class level.
    \item The instance-level prompt strategy leverages prompts that are tailored for individual instances in the private dataset, which are more informative for training models. We use BLIP-v2~\cite{li2023blip} to generate captions for each real image as the instance-level prompt (see Figure 2(b)). 
\end{itemize}
Once the client generates all the prompts, these prompts can be used as inputs to a specific pretrained text encoder (e.g., from CLIP~\cite{radford2021learning}) to generate all the corresponding embeddings. 

Basically, given a data point $(x, y)$, we first generate its corresponding prompt, denoted as $p$. Then, we create the text embedding $e$ based on the prompt $p$.
After receiving all $(e, y)$ from all clients, the server aggregates them for data synthesis with foundation generative models. By default, we use the Stable Diffusion as the generative model, and in Table~\ref{tb:gen}, we present the results obtained using different generative models. 


\subsection{Training Set Synthesis}
\label{sec:data generation}
After receiving all \emb{}, the server synthesizes every training sample $\bs_i$ by prompting the pre-trained Stable Diffusion with each $e_i$ as follows:
\setlength{\parskip}{0.8\parskip}
\begin{equation}
\bs_i = G(\bz_i, e_i) = \sqrt{\beta} \sum_{t=1}^{T} \sqrt{1-\beta^t} \cdot \frac{G_{\btheta_t}(\bz_i, e_i)}{\sqrt{T}} ,
\end{equation} 
where $\bz_i$ is a random noise vector, $e_i$ is the \emb{}, and $G_{\btheta_t}$ is the denoising network parameterized with $\btheta_t$ at time step $t$. The hyperparameter $\beta$ controls the trade-off between image quality and diversity, and $T$ is the number of diffusion steps. The inference process iteratively denoises the image then outputs the final synthetic training image. Finally, the server generates the synthetic training set $\ms = \{(\bs_i, y_i)\}_{i=1}^N$. Note that it is easy to synthesize more diverse training samples by combining multiple random noises with the sample prompt, thus training better models. In practice, we can adjust the number of synthetic training samples to trade-off the computational cost and performance.

\subsection{Model Updating}

\label{sec:modelupdating}
\subsubsection{One-shot Updating} 
We first show the efficacy of our approach in the context of one-shot federated learning~\cite{guha2019one,zhang2022dense}. This involves a central server learning a global model over a network of federated devices in a single round of communication.
Once the synthetic training set $\ms = \{(\bs_i, y_i)\}_{i=1}^N$ is obtained, we proceed to train a global model using only the synthetic data $\ms$ on the server. Subsequently, the trained model is sent to each client as the initial model.

\noindent \textbf{Robust to heterogeneous data and improved initial model.} Since collecting \emb{} during the initial round of communication is not affected by the varying data distribution. Therefore, our results remain consistent in different non-IID settings for one-shot FL.
In Figure~\ref{fig:acc_gap}, we have empirically demonstrated that FGL provides a well-trained initial model to each client at the first communication round. Consequently, in certain datasets, the performance of our method even exceeds that of central training after several rounds of communication (see Section~\ref{sec:imgnet_sub_exp} for the results).






\subsubsection{Multi-round Updating}
\label{sec:extend to multi-round communication}
Our method can also implement multi-round communication like traditional FL methods, which can bring further performance improvement. Since the one-shot results have already shown satisfactory performance, we assume that the multi-rounds of FGL with only 5 communication rounds will also be sufficient, as additional communication rounds are considered unnecessary. We further showcase the utility of server-side synthesized data in mitigating the forgetting problem that arises after model aggregation in highly non-IID scenarios~\cite{lee2021preservation,zhang2022federated}. Thus, for the multi-round FGL, we have implemented two versions: (1) directly utilizing the averaged model parameters without synthesized data; and (2) employing synthesized data for fine-tuning at the server-side, which is particularly useful for highly skewed data distributions. 
Note that the initial model is trained on the synthetic training set in the first-round communication for both two versions. 

\paragraph{Without Synthetic Data.} After the first round communication, each client receives the updated model from the server and then locally fine-tuning on it on private real training data, \ie $\btheta_k$. After locally fine-tuning, the server collects updated models from all clients for model aggregation, which is formulated as $\btheta = \frac{1}{K} \sum_{k=1}^{K} \btheta_k$, same as FedAvg. 
In other words, the server only aggregates models after the first round of communication.
This process is repeated until the model reaching the maximum communication rounds.
  \paragraph{With Synthetic Data.} 
   In the context of few-round communication scenarios, we find that training the aggregated model on synthesized data at the server-side can effectively mitigate the issue of forgetting~\cite{lee2021preservation} after aggregation, albeit at the expense of additional computational overhead. Specifically, fine-tuning the aggregated model on a synthetic training set in each communication round yields a substantial improvement in performance, particularly when dealing with highly imbalanced data distributions among clients.
More results are provided in 
Figure~\ref{fig:skewed}. 

\paragraph{Limitations of FGL.} FGL relies on powerful pretrained generative models. For FL tasks that have a similar data domains to the generative model's training data, FGL can yield decent performance in a single round. However, for very challenging data domains that may not be common in the generative model's training data, such as medical data, it requires a few more rounds to achieve satisfactory results. Further improvements can consider fine-tuning the generative model locally to better adapt to the specific data domain.

\section{Experimental Results}
\subsection{Experimental Setups}
\textbf{Data Partition}: Follow the setting in~\cite{fl_survey1,niidbench}, we adopt two data partition settings, namely, label distribution skew and feature distribution skew:
\begin{itemize}
    \item \textbf{Label Distribution Skew}:  Following in~\cite{zhang2022federated}, in which the label distributions varies on different clients, we employ the Dirichlet distribution $p\sim Dir(\beta)$ to simulate imbalanced label distributions. The hyper-parameter $\beta$ controls the degree of label imbalance, where a smaller value of $\beta$ indicates a more skewed label distribution.
    \item \textbf{Feature Distribution Skew}: In this setting, clients share the same label space while having a different feature distribution, which has been extensively studied in previous work~\cite{li2021fedbn,zhu2022aligning,yao2022federated,gong2022preserving}. We perform the classification task on natural images sourced from DomainNet~\cite{peng2019moment}, which consists of diverse distributions of natural images from six distinct data sources. 
\end{itemize}
\textbf{Baselines}: In our experiments, we select the popular FedAvg~\cite{fedavg} method and centralized training as the baselines by default\footnote{We also compared FGL with other baselines, \eg Moon~\cite{li2021model}, Fedopt~\cite{reddi2020adaptive}, as shown in Appendix Table~\ref{tb:baselines}.}. Assume that we have a total of 5 clients\footnote{Further results on 50 and 100 clients can be found in Table~\ref{tb:clients}. Varying the number of clients does not have a significant impact on our method, as long as the server can obtain all the text embeddings and subsequently synthesize well-generated data.} and that every client participates in communication. For both centralized training and federated learning, the local learning rate is set to 0.01, and we utilize the SGD optimizer with a momentum of 0.9. During the FedAvg training process, each client performs local updates for 5 epochs, and the communication round is set to 200. The centralized training consists of 120 rounds of iterations. 

\textbf{Implementation:} We employ the class-level prompt by default, which does not directly utilize local data information, thus providing enhanced privacy protection. We use Stable Diffusion v2-1-base model to construct synthetic data. 
We evaluate the performance of our method in scenarios of one-round (\ie \textbf{Ours (one-shot)}) and five-round communications (\ie \textbf{Ours (5-round)}). In the one-round communication scenario, we train the model for 120 epochs on the server. 
In the five-round communication scenario, we implement two variations:
(1) \textbf{Ours (5-round)}: based on the model trained in the first round, we perform four additional rounds of communication using the FedAvg algorithm.
(2) \textbf{Ours (5-round-syn)} : for extreme data distribution skew scenario, we further finetune the aggregated model using the synthetic dataset generated on the server during the first round for all five epochs. Please refer to \textbf{Appendix}~\ref{appendix:exp_settings} for more details on implementation and datasets.


\subsection{Results for Label Distribution Skew}
\label{sec:imgnet_sub_exp}

To evaluate the efficacy of our method on datasets with label distribution skew, 
we conduct experiments on five subsets of 224$\times$224 ImageNet~\cite{ILSVRC15}. Firstly, following \cite{howardsmaller,cazenavette2022dataset}, we do experiments on four datasets with 10 categories each, namely the coarse-grained ImageNette and ImageYellow, and the fine-grained ImageFruit and ImageSquawk.
We further conducted experiments on ImageNet100, involving 100 categories.
Also, we conducted experiments on two fine-grained image classification datasets, namely CUB-200~\cite{wah2011caltech}, Stanford Cars~\cite{krause20133d}.
We simulate three distinct dataset distributions, \ie IID, non-IID ($\beta=0.5$) and highly skewed distribution with $\beta=0.01$. 

The overall experimental results are shown in ~\cref{tb:imgnet_subset}. It is evident that our method with one-round communication outperforms the FedAvg method with 200 rounds of communication, by 6.0\%, 16.2\%, 7.8\% and 9.0\% on the four 10-category datasets in IID setting. Notably, our method is completely insensitive to data distribution in the first round. Hence, under extreme data distribution, \ie  $\beta=0.01$, our method surpasses FedAvg by 33.6\%, 42.8\%, 31.8\% and 39.2\% on the four 10-category ImageNet subsets respectively.
Also, our method outperforms FedAvg by a minimum of 30\% on ImageNet100, CUB-200, and Stanford Cars, after 5 rounds of communication, and even exceeds the performance of centralized training on these datasets. 

\begin{wrapfigure}{r}{0.45\textwidth}
\vspace{-7.6mm}
  \centering
  \includegraphics[width=0.44\textwidth]{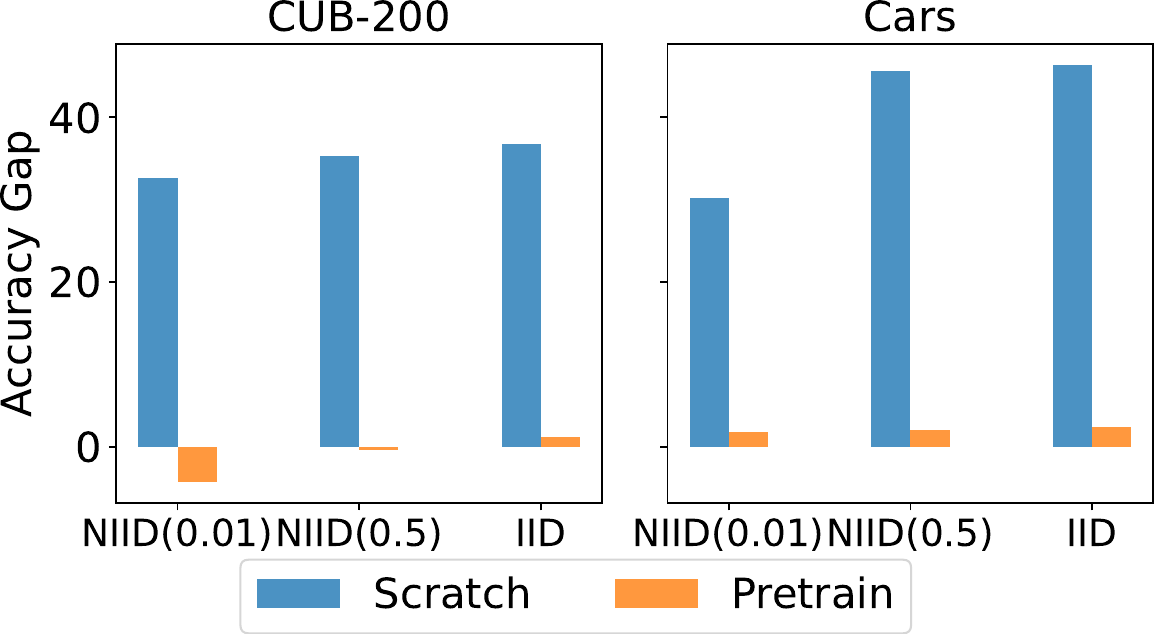}
  \vspace{-2mm}
  \caption{Accuracy gap when loading a pretrained model or training from scratch.}
  \label{fig:acc_gap}
  \vspace{-4mm}
\end{wrapfigure}

\paragraph{Why does training with synthetic data yield better results?}
\label{sec:why_works}
CUB-200 is a challenging dataset consisting of 200 bird species with 11,788 images, and Cars contains 16,185 images belonging to 196 classes of cars. The size of these fine-grained recognition datasets is typically smaller compared to general image classification datasets. In previous work~\cite{zhu2022dual,diao2022metaformer}, a common practice is to utilize a pretrained model that has been trained on the ImageNet dataset. In this study, we present two approaches: training the model from scratch and loading a pretrained ResNet34 model. Note that ImageNet has approximately 1.2M training data, but we only use about 0.18M images.
We present the accuracy gap between FGL and FedAvg in~\cref{fig:acc_gap}. It is evident that by directly loading a model pre-trained on ImageNet, the accuracy gap can be significantly reduced. 
\begin{table*}[t]
\centering
\vspace{-3mm}
  \caption{Performance comparison among different methods on 7 datasets. 
  The improvement $\textcolor{blue}{ _\uparrow }$ is compared to FedAvg IID results.}
\label{tb:imgnet_subset}
\scalebox{0.8}{
\begin{tabular}{c|ccc|c|ccc|c}
\toprule
{Dataset} & \multicolumn{3}{c|}{FedAvg}         & \multirow{2}{*}{Ours (one-shot)}  & \multicolumn{3}{c|}{Ours (5-round)}  & \multirow{2}{*}{Centralized}         \\
                         & $\beta=0.01$ & $\beta=0.5$ & IID   &                                  & $\beta=0.01$ & $\beta=0.5$ & {IID} &                                 \\
\midrule
ImageNette               & 51.6         & 75.0          & \underline{79.2}  & $85.2_{\textcolor{blue}{ \uparrow  6.0}}$                                                    & 82.8         & 94.0          & $\textbf{95.6}_{\textcolor{blue}{ \uparrow  16.4}}$  & 92.2 \\
ImageFruit               & 29.0           & 51.2        & \underline{55.6}  & $71.8_{\textcolor{blue}{ \uparrow 16.2}}$                                                     & 67.2         & 80.2        & $\textbf{83.2}_{\textcolor{blue}{ \uparrow  27.6}}$   & 78.2 \\
ImageYellow              & 50.6         & 70.2        & \underline{74.6}  & $82.4_{\textcolor{blue}{ \uparrow 7.8}}$                                                      & 79.4         & 91.0          & $\textbf{94.8}_{\textcolor{blue}{ \uparrow  20.2}}$  & 90.8 \\
ImageSquawk              & 49.6         & 73.2        & \underline{79.8}  & $88.8_{\textcolor{blue}{ \uparrow 9.0}}$                                                      & 90.0           & 95.0          & $\textbf{95.6}_{\textcolor{blue}{ \uparrow  15.8}}$  & 92.4 \\ \midrule
ImageNet100              & 36.3        & 44.6        & \underline{49.4} & $48.4_{\textcolor{blue}{ \downarrow 1.0}}$                                                     & 70.1         & 74.9       & $\textbf{80.1}_{\textcolor{blue}{ \uparrow  30.7}}$  & 77.0 \\
\midrule
CUB-200              & 35.0         & 36.6        & \underline{36.6}  & $44.6_{\textcolor{blue}{ \uparrow 8.0}}$                                                      & 67.7         & 71.9         & $\textbf{73.3}_{\textcolor{blue}{ \uparrow 37.3 }}$  & 48.3 \\
\midrule
Stanford Cars              & -         & 42.4       & \underline{44.5}  & $54.23_{\textcolor{blue}{ \uparrow 9.7}}$                                                      & 85.4         & 88.0         & $\textbf{88.8}_{\textcolor{blue}{ \uparrow 44.3 }}$  & 64.7 \\
\bottomrule
\end{tabular}}
\vspace{-5mm}
\end{table*}

\textit{\textbf{Takeaways}}:
In the first round of communication, FGL generates a set of synthetic data to train a model, which serves as an excellent initial model. Unlike `out-of-domain' datasets such as ImageNet, this smaller `in-domain' synthetic data exhibits remarkable performance in FL tasks.

\begin{wrapfigure}{r}{0.45\textwidth}
\vspace{-2.8mm}
  \centering
  \includegraphics[width=0.44\textwidth]{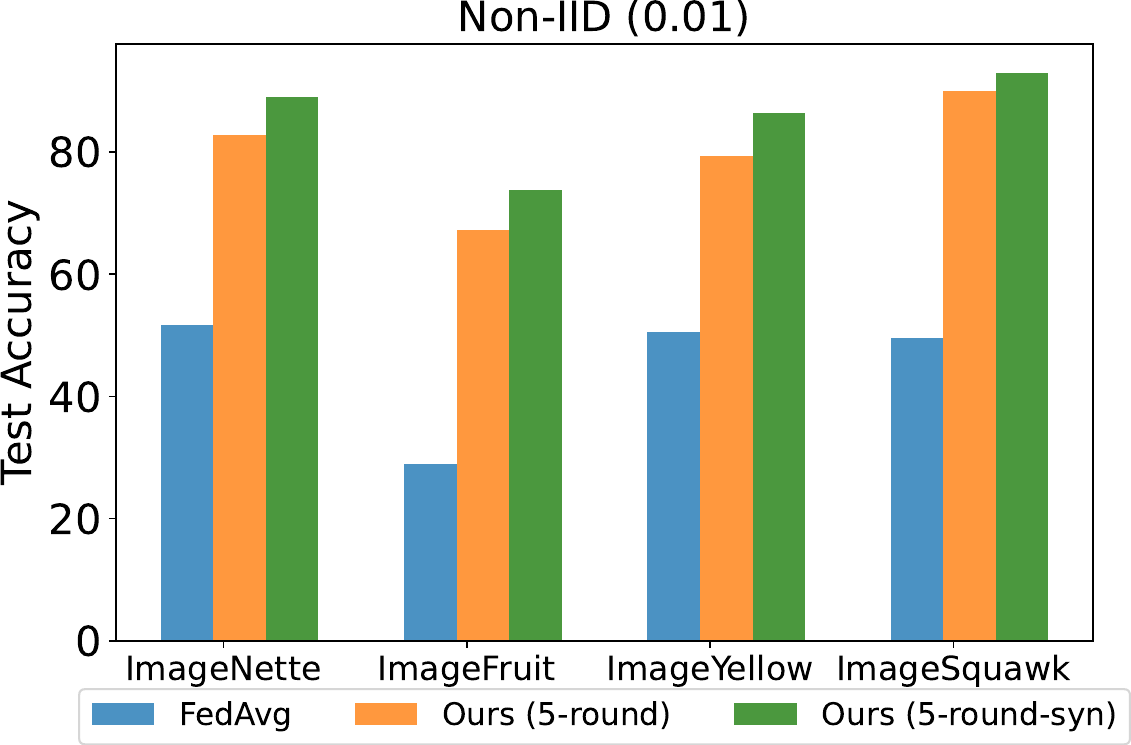}
  \vspace{-2mm}
  \caption{Accuracy on highly skewed data.}
  \label{fig:skewed}
  \vspace{-3.5mm}
\end{wrapfigure}
\paragraph{Results on Highly Skewed Data.} Table~\ref{tb:imgnet_subset} shows that when the data distribution is extremely skewed ($\beta=0.01$), Ours (5-round) does not outperform Ours (one-shot) on Imagenette, ImageFruit, and ImageYellow datasets. We attribute this phenomenon to the fact that in case of highly skewed data distribution, more rounds of communication are required for the models to converge gradually~\cite{zhao2018federated,wang2020tackling}. To achieve better results within five rounds, we perform fine-tuning on the aggregated model using the synthesized dataset from the first round on the server for 5 epochs before distributing the updated model. The results are presented in~\cref{fig:skewed}, which clearly demonstrate that fine-tuning on our synthesized dataset can significantly enhance model performance even with extreme data distribution.

\begin{wrapfigure}{r}{0.45\textwidth}
\vspace{-4.8mm}
  \centering
\includegraphics[width=0.38\textwidth]{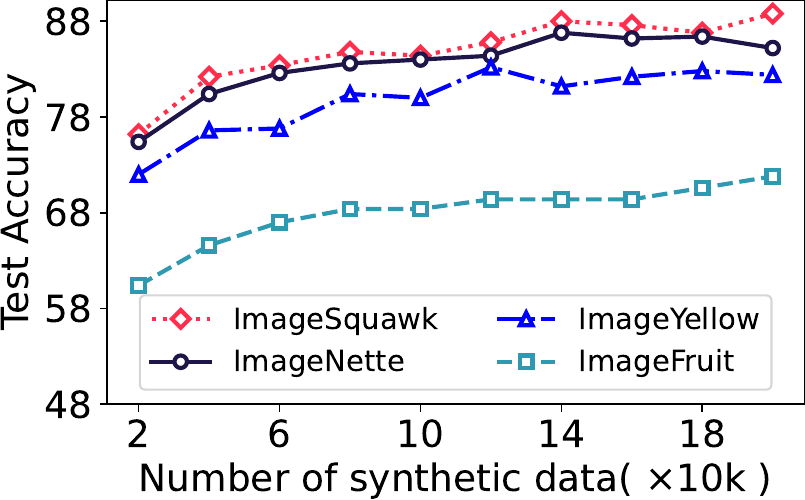}
  \vspace{-3mm}
  \caption{Varying synthetic data volume.}
  \label{fig:num_syn}
  \vspace{-3mm}
\end{wrapfigure}
\paragraph{Number of Synthetic Data.} 
By comparing the results of 1.3$k$ synthetic images per class in Appendix (Fig.~\ref{fig:blip}) and 20$k$ synthetic images per class~\cref{tb:imgnet_subset}, we find the performance of our method can significantly improve by synthesizing more training samples, \ie from 73.2\% to 85.2\% on the Imagenette dataset. We further study the influence of synthetic image number and model performance in~\cref{fig:num_syn} on four 10-category datasets. We synthesize more images per class by integrating the prompts and more random noises. Obviously, as the number of images per class increases from 2k to 20k, the test accuracy improves consistently.

\begin{wrapfigure}{r}{0.45\textwidth}
\vspace{-6mm}
  \centering
  \includegraphics[width=0.4\textwidth]{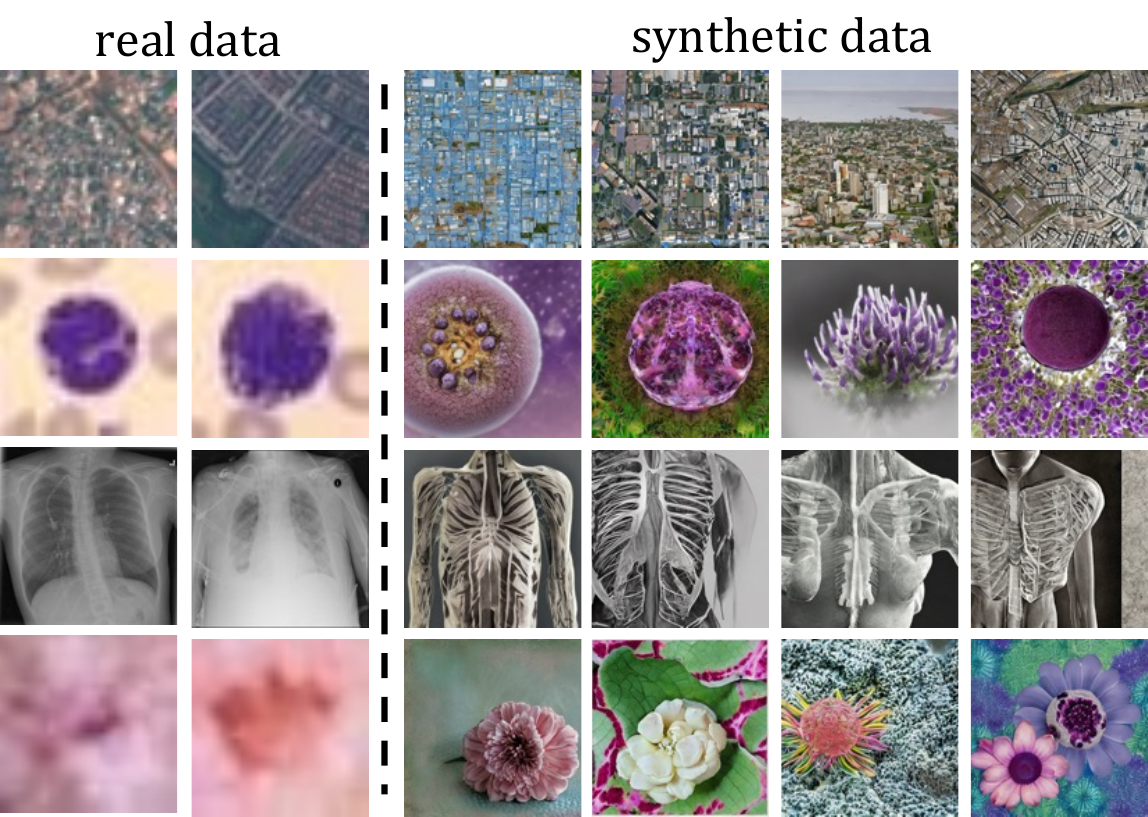}
  \vspace{-2mm}
  \caption{Visualization of real data and synthetic data, where the datasets EuroSAT, BloodMNIST, COVID-19 X-rays, and DermaMNIST are displayed from top to bottom.}
  \label{fig:hard_domain_show}
  \vspace{-4mm}
\end{wrapfigure}
\paragraph{Results on Medical and Satellite Datasets}
We further show the performance of our method on datasets with challenging domains such as medical datasets and satellite images to explore the limits of our approach. ~\cref{fig:intro_b} illustrates the performance of FedAvg and our proposed method on the EuroSAT~\cite{helber2019eurosat} satellite dataset as well as three medical datasets, namely COVID-19 X-rays~\cite{chowdhury2020can}, BloodMNIST and DermaMNIST~\cite{medmnistv2}. 
Our method demonstrates comparable performance to FedAvg (200-round) after 5 communication rounds on both medical and satellite datasets.
~\cref{fig:hard_domain_show} presents the visualization of the synthetic and real data, highlighting the difficulties encountered in generating data for these challenging domains. The capability of the generative model is somewhat diminished in these scenarios. Please refer to the Appendix (Table~\ref{tb:chall}) for more detailed results under different non-IID settings.

\textbf{\textit{Takeaways}}:
The quality of synthetic data may be influenced by the domain discrepancy between the local training data and the pretraining data used for the foundation model. This discrepancy ultimately limits the performance of FGL. One potential solution is to finetune the foundation model using local data, however, this could potentially raise privacy concerns. We leave it for future research.

\subsection{Results for Feature Distribution Skew }
To simulate the scenario of feature distribution skew, we select 10 categories from the DomainNet dataset to conduct experiments . Each client is assigned a specific domain, and we have a total of 6 clients participating in FL. To ensure an adequate amount of data, we synthesize 3,500 samples for each class within each domain, resulting in a cumulative dataset of 210$k$ samples.
~\cref{tb:domainnet} presents the performance of various methods on six domains respectively and their average accuracy. It shows that in the one-shot communication scenario, our method outperforms FedAvg by 2\% to 19\% in five domains, but exhibits notably poor performance in the Quickdraw domain. To investigate the underlying reason, we visualize the synthetic and real data in {Appendix}~\ref{appendix:exp_1}. 
It becomes apparent that this performance decline is attributed to the difficulty for diffusion model to synthesize images that align with Quickdraw domain when using the class-level prompt, \ie  ``A black and white drawing of a \{class name\}''. We present the results of the generative model on medical and satellite datasets in Figure~\ref{fig:hard_domain_show}, which also demonstrate the limited synthetic capability of the generative model in challenging domains.

However, when implementing a five-round communication experiment, our method demonstrates a 2.2\% performance improvement over FedAvg specifically on the Quickdraw domain, and an overall performance improvement of 11.75\%. Interestingly, our method even surpasses the performance of the centralized training models by 5.95\%. We provide further results on each domain in Appendix~\ref{appendix:exp_2}, and more visualizations of synthetic data on DomainNet and ImageNet are shown in Appendix~\ref{appendix:exp_3}.


\begin{table*}[t]
\vspace{-2mm} 
\centering
\caption{Performance for feature distribution skew. Each client hosts data from a specific domain of DomainNet dataset.}
\label{tb:domainnet}
\scalebox{0.82}{
\begin{tabular}{c|cccccc|c}    
\toprule
Method                      & Clipart        & Infograph      & Painting       & Quickdraw      & Real           & Sketch        & Average            \\
\midrule
FedAvg                      & \underline{80.97}          & \underline{41.90}          & \underline{57.33}         & \underline{78.93}          & \underline{80.56}          & \underline{70.06}         & \underline{72.30 }          \\
Centralized                 & 81.37          & 50.82          & 60.63          & 92.46          & 82.20          & 73.93         & 78.10         \\
\midrule
Ours (one-shot)             & $83.40_{\textcolor{blue}{ \uparrow  2.43}}$          & $49.58_{\textcolor{blue}{ \uparrow 7.68}}$          & $76.88_{\textcolor{blue}{ \uparrow 19.55}}$          & $51.80_{\textcolor{blue}{ \downarrow 27.13}}$          & $87.06_{\textcolor{blue}{ \uparrow 6.5}}$          & $81.10_{\textcolor{blue}{ \uparrow 11.04}}$         & $71.59_{\textcolor{blue}{ \downarrow 0.71}}$         \\
\textbf{Ours (5-round)} & $\textbf{90.89}_{\textcolor{blue}{ \uparrow 9.92}}$ & $\textbf{61.61}_{\textcolor{blue}{ \uparrow 19.71}}$ & $\textbf{79.52}_{\textcolor{blue}{ \uparrow 22.19}}$ & $\textbf{81.13}_{\textcolor{blue}{ \uparrow 2.2}}$ & $\textbf{91.13}_{\textcolor{blue}{ \uparrow 10.57}}$ & $\textbf{90.20}_{\textcolor{blue}{ \uparrow 20.14}}$ & $\textbf{84.05}_{\textcolor{blue}{ \uparrow 11.75}}$ \\
\bottomrule
\end{tabular}}
\vspace{-6mm}
\end{table*}


\subsection{Privacy Analysis}
\label{sec:privacy}

\paragraph{Membership Inference Attack (MIA).} 
\begin{wrapfigure}{r}{0.4\textwidth}
\vspace{-6mm}
  \centering
  \includegraphics[width=0.38\textwidth]{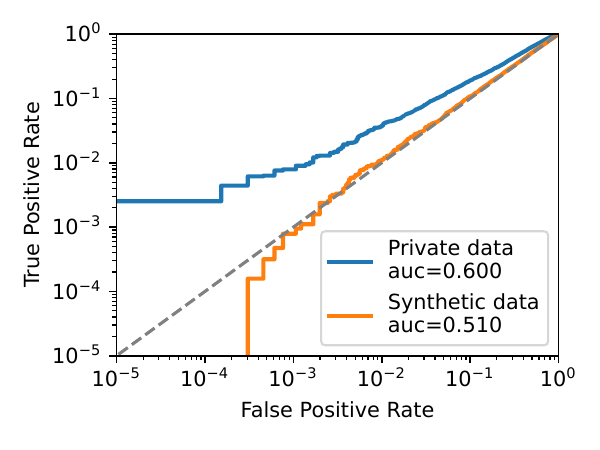}
  \vspace{-4mm}
  \caption{Attack results under LiRA for models trained on real data and synthetic data. A total of 32 shadow models were trained, ensuring that each sample was trained on half of these models.}
  \label{fig:lira}
  \vspace{-3mm}
\end{wrapfigure}

The objective of MIA is to examine whether a specific data point belongs to the training set used to train a machine learning model. Given that Ours (one-shot) does not depend on real training data, it is reasonable to assume that it may not encounter any privacy leakage. However, since Ours (5-round) and FedAvg both utilize private data for training, we present the MIA results on ImageNette using the low false-positive rate regime, as recommended by the state-of-the-art Likelihood Ratio Attack (LiRA)~\cite{carlini2022membership}. 
As show in Figure~\ref{fig:lira}, when employing the LiRA against models trained using private data (\ie FedAvg) and synthetic data (\ie Ours (5-round)), the latter exhibits a stronger defense against membership inference attacks. This can be attributed to the fact that our model, trained on synthetic data, exhibits minimal information leakage.

\paragraph{Detecting Content Replication and Memorization.}
\begin{wrapfigure}{r}{0.4\textwidth}
\vspace{-6mm}
  \centering
  \includegraphics[width=0.38\textwidth]{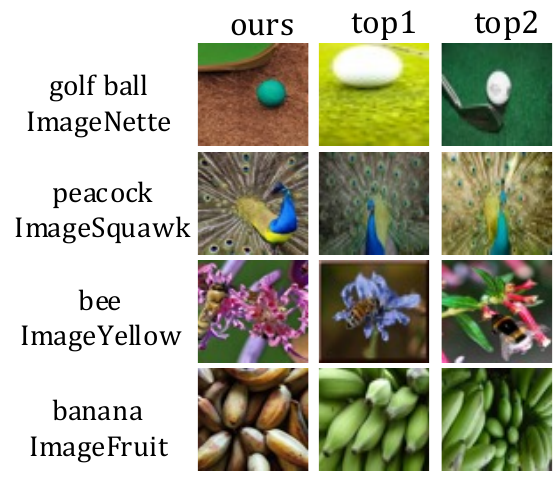}
  \caption{Retrieving similar real images for each synthetic image.}
  \label{fig:memorization}
  \vspace{-4mm}
\end{wrapfigure}

Previous research~\cite{carlini2023extracting,zhang2023forget,van2021memorization,somepalli2023understanding,somepalli2023diffusion} has indicated that diffusion models store and reproduce specific images from their training dataset during the generation process.
Although our training process (with class-level prompts) does not access the private data of clients, we still discuss the potential privacy risks that may arise. 
Follow the setting in~\cite{somepalli2023diffusion}, we conduct image retrieval experiments, which allows us to compare the synthetic images with the original training images and detect any instances of content duplication. We perform a quantitative analysis on 1000 synthetic images across four datasets. For each synthetic image, we search the training set by computing the Cosine similarity between its feature and features of real training images. Figure~\ref{fig:memorization} showcases the top 2 most similar images from each of the four datasets. 
These images exhibit no noteworthy similarities in background and foreground, which verifies FGL doesn't compromise privacy of the client's private data.

\subsection{Ablation Study}
\paragraph{Varying the Generative Models.} To investigate the impact of various generative models on the results, we followed the setting in~\cite{li2023synthetic}. Our experiments primarily focus on three prevalent conditional diffusion models: DiT~\cite{dit}, GLIDE~\cite{glide}, and Stable Diffusion. We use these off-the-shelf models to generate synthetic images. Specifically, for GLIDE and Stable Diffusion, the prompt was configured as ``a photo of \{label name\}, real-world images, high resolution''. For DiT, the input comprised the label ID corresponding to the ImageNet1k dataset. The images synthesized by DiT and GLIDE are of dimensions 256x256, whereas those produced by Stable Diffusion are of dimensions 512x512. As shown in Table~\ref{tb:gen}, even when we vary the foundation models used in our method, FGL consistently outperforms FedAvg by a significant margin. 

\paragraph{Results on more clients and more baselines} To demonstrate the scalability of our method to a larger number of clients, we extended our analysis to include the results obtained from the ImageNette dataset with 50 and 100 clients. As depicted in Table~\ref{tb:clients} (see the appendix), our method continues to exhibit superior performance compared to FedAvg across all scenarios. Additionally, the improvements achieved by our method remain significant. 
Also, we have compared the two popular FL methods, Moon~\cite{li2021model} and Fedopt~\cite{reddi2020adaptive}. We conducted experiments on the ImageNette and ImageNet100 datasets, considering a scenario with 50 clients under non-IID settings ($\beta=0.5$). As shown in the Table~\ref{tb:baselines} (see the appendix), our method still outperforms other FL approaches.

\begin{table}[t]
\centering
\vspace{-6mm}
\caption{Ablation study on the generative model used in FGL.}
\label{tb:gen}
\scalebox{0.8}{
\begin{tabular}{c|c|c|c|c|c}
\toprule
Method                    & one-shot & \begin{tabular}[c]{@{}c@{}}5-round, \\ $\beta=0.01$\end{tabular}  & \begin{tabular}[c]{@{}c@{}}5-round, \\ $\beta=0.5$\end{tabular}  & IID  & Centralized \\
\midrule
Ours w/  SD & \textbf{85.2}     & \textbf{82.8}               & \textbf{94.1}                & \textbf{95.6} & 92.2        \\
Ours w/ Glide             & 79.0       & 76.2               & 89.4              & 89.4 & 92.2        \\
Ours w/ Dit               & 76.2     & 74.6               & 90.2              & 92.8 & 92.2        \\
\midrule
FedAvg (120-round)        & -        & 51.6               & 75.1              & 79.2 & 92.2     \\
\bottomrule
\end{tabular}}
\vspace{-6mm}
\end{table}

\vspace{-3mm}
\section{Related Work}
\paragraph{Foundation Generative Models.} Large generative models, such as Stable Diffusion ~\cite{rombach2022high}, DALL-E2 ~\cite{ramesh2022hierarchical}, Imagen ~\cite{saharia2022photorealistic}, and GLIDE ~\cite{nichol2021glide}, have recently emerged as an off-the-shelf tool for high-quality and real-looking image generation conditioned on text prompts. 
A few works have explored the usage of synthetic images as training data.
For example, He et al. ~\cite{he2022synthetic} show that synthetic data generated by diffusion models can improve pretraining, zero-shot, and few-shot image classification performance. Li ~\cite{li2023synthetic} demonstrate that synthetic data generated by conditional diffusion models can be used for knowledge distillation without original data.
Zhou~\cite{zhou2023training} synthesize better images for model training with stable diffusion by implementing diffusion inversion. 
\paragraph{Foundation Models in FL.} The foundation generative models are still under-explored in federated learning, though there exist a few related works that study foundation models in federated learning. 
The most similar one is \cite{yang2023exploring}, which takes advantage of the diffusion model in the server to synthesize training samples that complied to the distributions of domain-specific features from clients.
Yu et al. \cite{yu2023federated} introduce federated learning into foundation model training for training foundation models collaboratively with private data. Like traditional FL methods, they also transmit model parameters between servers and clients. Based on the shared CLIP model, Guo et al. \cite{guo2022promptfl} transmit the small number of updated parameters of the prompt learner from clients to the server to reduce the communication cost. 
\label{sec:preliminary}



\section{Conclusion}
In this work, we introduce a pioneering framework, named \emph{Federated Generative Learning}, which transmits prompts associated with distributed training data between clients and the server. By leveraging foundation generative models, informative training data can be synthesized remotely using received prompts that contain minimal privacy. The proposed framework exhibits several noteworthy advantages, including improved communication efficiency, better resilience to distribution shift, substantial performance gains, and enhanced privacy protection. 

\bibliographystyle{plain}
\bibliography{ref}
\newpage

\section{Appendix}
\begin{table}[]
\centering
\caption{Results on more clients.}
\label{tb:clients}
\scalebox{0.9}{
\begin{tabular}{c|cc|c|cc|c}
\toprule
 & \multicolumn{2}{c|}{FedAvg} & Ours (one-shot) & \multicolumn{2}{c|}{Ours (5-round)} & Centralized \\
Client & $\beta=0.5$ & FedAvg (IID) &      & $\beta=0.5$ & IID  &      \\
\midrule
5      & 75          & 79.2         & 85.2 & 94          & 95.6 & 92.2 \\
50     & 72          & 77.0           & 85.2 & 93.8        & 91.2 & 92.2 \\
100    & 70.0          & 67.2         & 85.2 & 92.8        & 93.2 & 92.2 \\
\bottomrule
\end{tabular}}
\end{table}

\begin{table*}[t]
\centering
\caption{Examples of prompt generation patterns for Class-Level and Instance-Level prompting}
\label{tab:prompt_setting}
\scalebox{0.82}{
\begin{tabular}{c|c|c|c}
\toprule
\textbf{Prompt Type} & \textbf{Dataset} & \textbf{Prompt Template} & \textbf{Prompt Example} \\
\midrule
Class Level & ImageNet Subset & label + `, real world images & tench, real world images, \\
            &                & , high resolution' & high resolution \\
\midrule
Class Level & DomainNet Subset & label + style & an airplane, Sketch drawing \\
            &               & & with only one object in the picture \\
\midrule
Instance Level & ImageNet Subset & label + `, ' + image caption + `, & tench, Tinca tinca, a man  \\
 &  & real world images, high resolution.' & kneeling down holding a fish in the grass \\
\bottomrule
\end{tabular}}
\end{table*}
\begin{table*}[]

\centering
  \caption{Detailed description of the DomainNet Subset Dataset}
  \label{tab:domainNet_detail}
  \scalebox{0.9}{
\begin{tabular}{@{}c|c|c|c@{}}
\toprule
Description &
  \# class &
  Class name &
  Domain name \\ \midrule
10 classes from DomainNet &
  10 &
  \begin{tabular}[c]{@{}c@{}}airplane, clock, axe, \\ basketball, bicycle, bird, \\ strawberry, flower, pizza, bracelet\end{tabular} &
  \begin{tabular}[c]{@{}c@{}}Clipart, Infograph, Painting,\\  Quickdraw, Real, Sketch\end{tabular} \\ \bottomrule
\end{tabular}}
\end{table*}

\begin{table*}[]
\centering
\caption{Detailed description of the ImageNet Subset Dataset}
\label{tab:imageNet_detail}
\scalebox{0.92}{
\begin{tabular}{c|c|c|c|c}
\toprule
\textbf{Dataset} & \textbf{Description} & \textbf{\# Class} & \textbf{Class name} & \textbf{Class id} \\
\midrule
lmageNette & 10 class & 10 & (tench, English springer, & (0, 217, 482, 491, 497, \\
 & from ImageNet &  & cassetteplayer, chain saw,  &  566, 569,571, 574, 701) \\
 &  &  & church, Frenchhorn, & \\
 &  &  &  garbage truck, gas pump, &  \\
 &  &  &  golfball, parachute)& \\
\midrule
 
lmageFruit & 10 class & 10 & (pineapple, banana,  & (953, 954, 949, 950, 951, \\
 & from ImageNet &  & strawberry, orange,  &  957, 952, 945, 943, 948) \\
 &  &  &lemon, pomegranate, & \\
 &  &  &  fig, bell pepper,  &  \\
 &  &  &  cucumber, green apple) & \\
\midrule

lmageYellow & 10 class & 10 & (bee, ladys slipper,  & (309,986, 954, 951, 987, \\
 & from ImageNet &  & banana, lemon,  &  779, 599, 291, 72, 11) \\
 &  &  & corn,  school bus, & \\
 &  &  &  honeycomb,  lion, &  \\
 &  &  &  garden spider, goldfinch) & \\
\midrule

lmageSquawk & 10 class & 10 & (peacock, flamingo, & (84, 130, 88, 144, 145, \\
 & from ImageNet &  & macaw, pelican,  &   22, 96, 9, 100, 89) \\
 &  &  & king penguin, bald eagle, & \\
 &  &  & toucan, ostrich, &  \\
 &  &  &  black swan, cockatoo)& \\
\midrule

ImageNet100 & 100 class & 100 & - - &- - \\
 & from ImageNet &  & &   \\

\bottomrule
\end{tabular}}
\end{table*}

\subsection{Experiments Setting}
\label{appendix:exp_settings}
\subsubsection{Prompt Generation for Synthetic Data in ImageNet and DomainNet}
In this section, we present the configurations for prompt generation when synthesizing data for the ImageNet and DomainNet datasets. As depicted in Table ~\ref{tab:prompt_setting}, for class-level prompt generation on ImageNet-like datasets, the prompt template consists of the label followed by ``, real-world images, high resolution.'' Examples of generated prompts include ``tench, real world images, high resolution'' and ``English springer, real world images, high resolution.'' On DomainNet Subset datasets, the prompt template comprises the label and style, where the label represents the category name, and the style describes the domain. Examples of generated prompts in this context are ``an airplane, Sketch drawing with only one object in the picture'' and ``an airplane, real world images, high resolution, with only one object in the picture.''

For instance-level prompt generation on ImageNet Subset datasets, the prompt template consists of the label followed by ``, '' and the image caption followed by ``, real-world images, high resolution.'' Here, the label represents the category name, and image caption corresponds to the textual description of the image generated by BLIPv2. Examples of generated prompts include ``tench, Tinca tinca, a man kneeling down holding a fish in the grass'' and ``tench, Tinca tinca, a man kneeling down holding a large fish in the water.''

\subsubsection{Dataset Description}
In this section, we provide a detailed description of the datasets used in our experiments. The datasets include the DomainNet Subset, lmageNette, lmageFruit, lmageYellow, lmageSquawk, ImageNet100, Eurosat, COVID-19 X-rays, BloodMNIST and DermaMNIST.

\textbf{DomainNet Subset}: This subset is selected from the DomainNet dataset and consists of ten categories spanning six different domains. Refer to Table ~\ref{tab:domainNet_detail} for detailed class and domain names.

\textbf{ImageNet Subset}: These datasets are subsets extracted from ImageNet, including lmageNette, lmageFruit, lmageYellow, lmageSquawk, and ImageNet100. Refer to Table  ~\ref{tab:imageNet_detail} for details on class names and class IDs.

\textbf{Eurosat}: The Eurosat dataset~\cite{helber2019eurosat}, derived from Sentinel-2 satellite images with 13 spectral bands, consists of 10 classes, totaling 27,000 labeled and geo-referenced images. Official images are of size \(64 \times 64\), whereas our synthesized data is \(512 \times 512\). To enhance diversity, we randomly crop the synthesized data to \(64 \times 64\), \(128 \times 128\), or \(224 \times 224\), followed by resizing to \(64 \times 64\) for training. 

\textbf{COVID-19 X-rays}: The COVID-19 X-rays dataset~\cite{chowdhury2020can}, categorized into COVID-19, normal, and viral pneumonia classes, follows the COVIDx-8A version with 5,585 training images and 400 test images. During training, images are resized to \(256 \times 256\) and subjected to random resized cropping to \(224 \times 224\). 

\textbf{MedMNIST v2 Subset}: Both the BloodMNIST and DermaMNIST datasets are constituents of MedMNIST v2~\cite{medmnistv2}, a comprehensive MNIST-like compilation of standardized biomedical images.The DermaMNIST dataset originates from HAM10000, a vast collection of dermatoscopic images of common pigmented skin lesions from multiple sources. The dataset comprises 10,015 dermatoscopic images categorized into seven different diseases, structured as a multi-class classification task. We follow the official partitioning of training, validation, and test sets with a ratio of 7:1:2. The image dimensions are uniform at $3 \times 28 \times 28$, and for synthesized data, we resize to $3 \times 28 \times 28$ for training.The BloodMNIST dataset is derived from individual normal cells captured from individuals devoid of infection, hematologic or oncologic diseases, and without any pharmacologic treatment at the time of blood collection. Comprising a total of 17,092 images distributed across eight classes, we adopt the official partitioning with a ratio of 7:1:2 for training, validation, and test sets. The image dimensions are consistently $3 \times 28 \times 28$, and for synthesized data, we resize to $3 \times 28 \times 28$ during training.


\subsection{Additional Experiments}

\subsubsection{Synthetic Data Visualization on Quickdraw Domain}
\label{appendix:exp_1}
\begin{figure*}[h] 
    \centering
    \includegraphics[width=14cm]{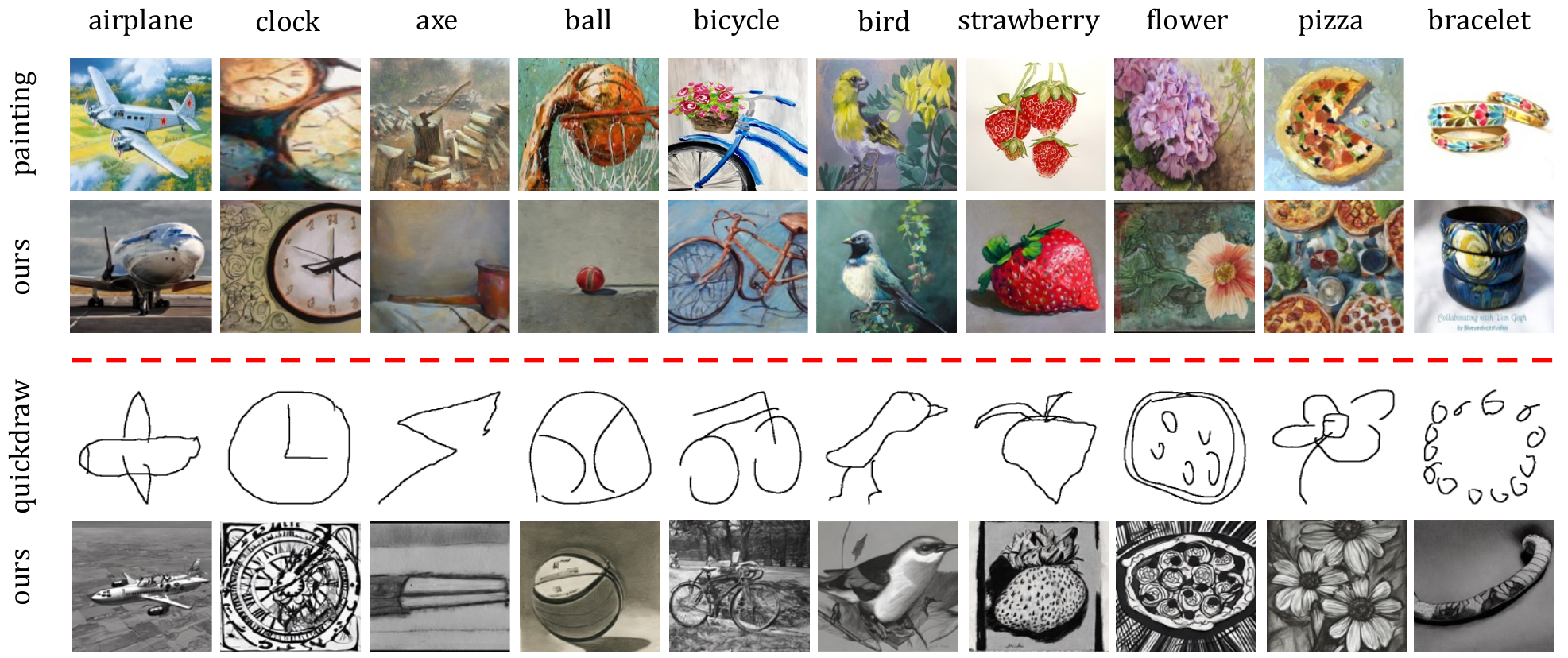}
 \caption{Visualization of original and synthetic data on the Painting and QuickDraw domains of DomainNet dataset. Obviously, it is easy for diffusion model to synthesize painting-like images, while challenging to synthesize images with QuickDraw style. }
    \label{fig:domainnet_view}
\end{figure*}

To investigate the reasons behind the notably poor performance in the Quickdraw domain, we visualize both synthetic and real data for the Painting and QuickDraw domains of the DomainNet dataset in Figure \ref{fig:domainnet_view}. The first and second rows correspond to real data and synthetic data for the Painting Domain, respectively. It is evident that the synthetic data closely resembles the real data in style, which results in the model performing well in this domain. The third and fourth rows represent real data and synthetic data for the Quickdraw Domain, respectively. Notably, there is a substantial stylistic difference between the synthetic data and real data in this domain, providing an explanation for the poor model performance.

\subsubsection{Testing Accuracy on Each Domain.} 
\label{appendix:exp_2}
\begin{figure*}[] 
    \centering
    \includegraphics[width=14cm]{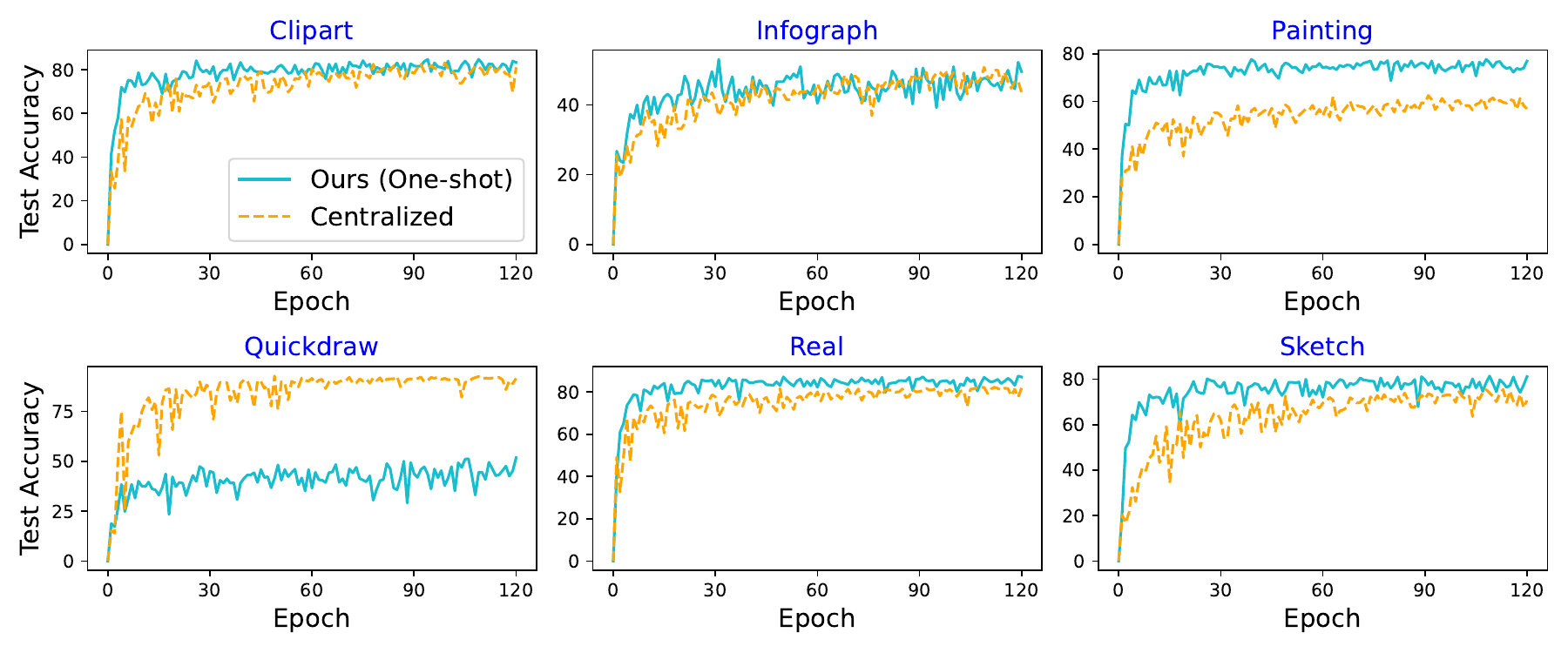}
 \caption{Performance comparison between Ours (One-shot) and the Centralized Training on six domains of DomainNet dataset.}
    \label{fig:domainnet_acc}
\end{figure*}

In this section, we compare the performance of the ``Centralized'' and ``Ours (One-shot)'' methods on the DomainNet dataset across different domains during the training process. The experimental results, as shown in Figure~\ref{fig:domainnet_acc}, demonstrate that, except for the ``Quickdraw'' domain, Ours (One-shot) outperforms the Centralized Training in five domains. Particularly, in the ``Painting'' domain, our method achieves a significant performance improvement.
We provide further explanation about the performance gap in ``Quickdraw'' domain through visualization in the following section. 

\begin{figure*}[] 
    \centering
    \includegraphics[width=16cm]{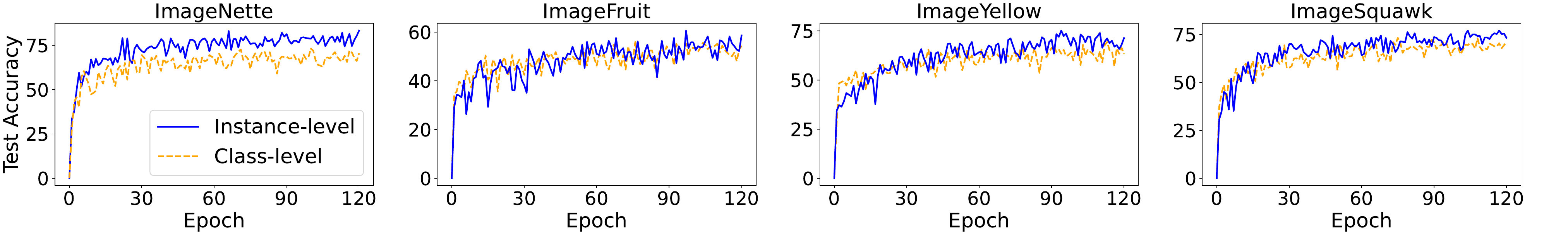}
  \caption{Accuracy of class-level prompt and instance-level prompt on four datasets. We synthesize 1300 images per class. After using instance-level prompts, the best accuracy improved by 10.2\%, 4.6\%, 5.8\%, and 3.8\% respectively.}
    \label{fig:blip}
\end{figure*}

\subsubsection{Visualization on DomainNet and ImageNet}
\label{appendix:exp_3}
\begin{figure*}[t] 
    \centering
    \includegraphics[width=14cm]{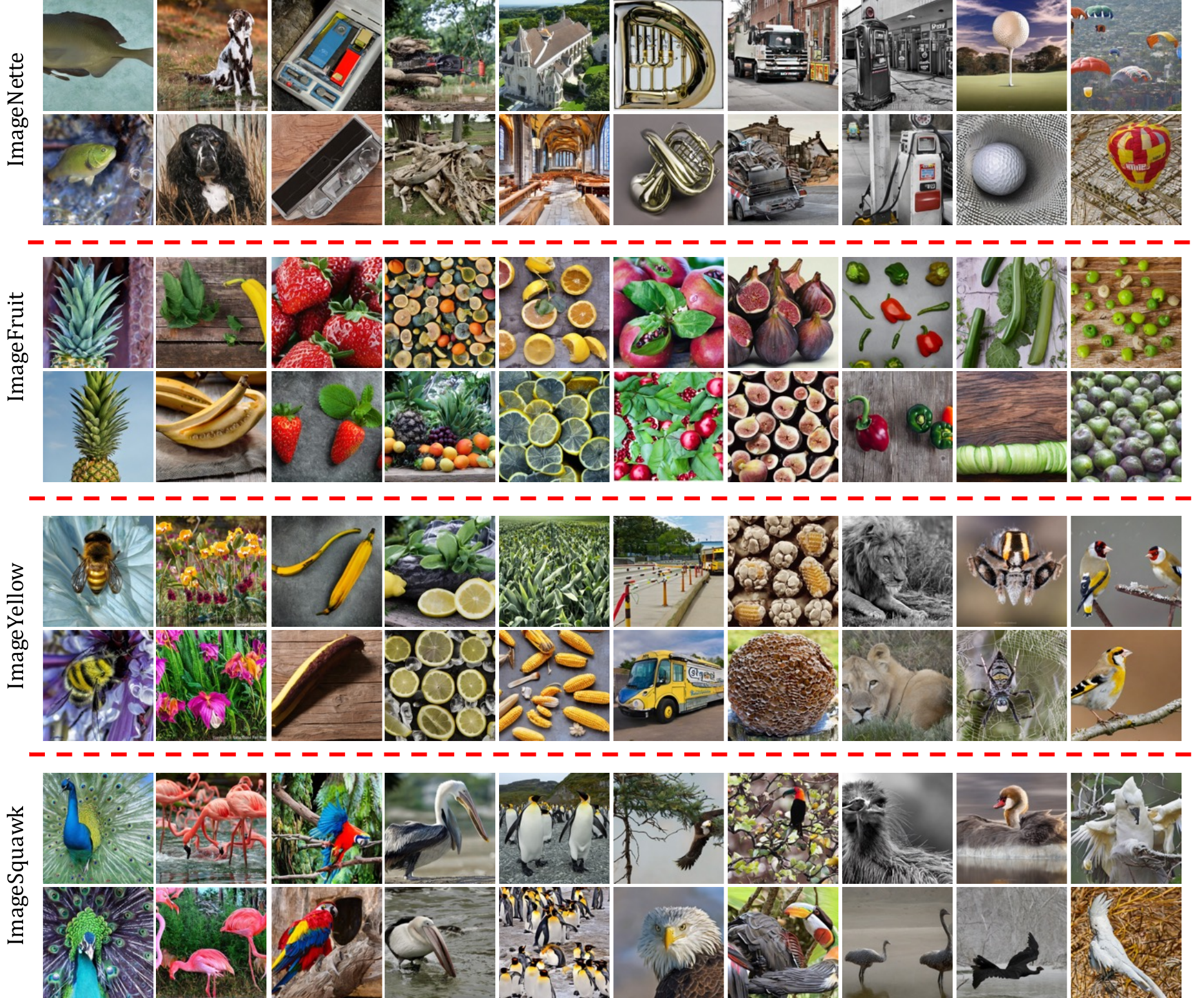}
 \caption{Visualization of synthetic images of four ImageNet subsets.}
    \label{fig:imagenet_visual}
\end{figure*}

\begin{figure*}[] 
    \centering
    \includegraphics[width=14cm]{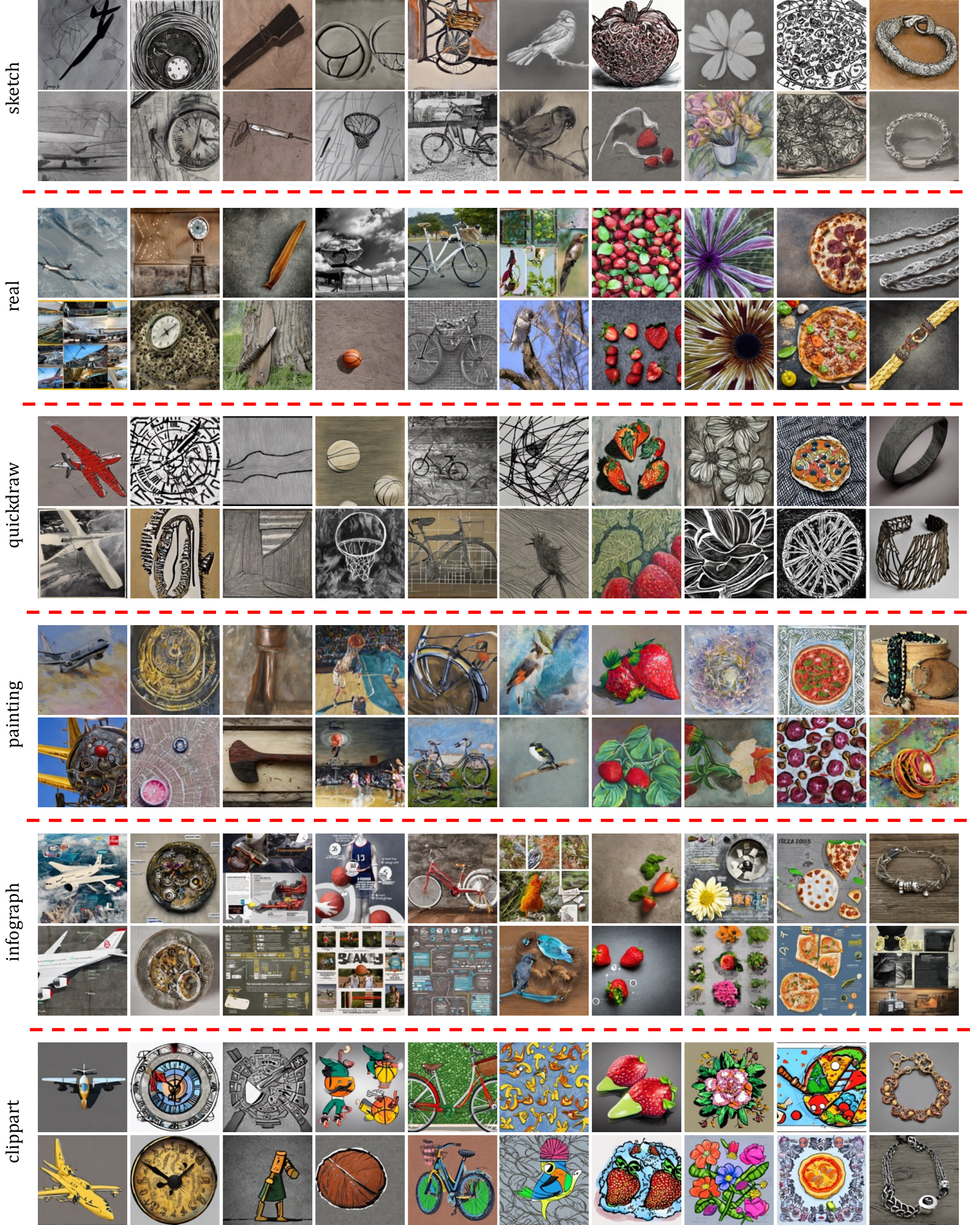}
 \caption{Visualization of synthetic images of six domains from DomainNet dataset.}
    \label{fig:domainnet_visual}
\end{figure*}

In this section, we provide comprehensive visualizations of the synthetic data generated from the ImageNet and DomainNet datasets. Figure \ref{fig:imagenet_visual} showcases visualizations of synthetic data from four distinct subsets of the ImageNet dataset. Each pair of rows corresponds to one of these subsets, which includes ImageNette, ImageFruit, ImageYellow, and ImageSquawk subdatasets. Within each column, individual images represent specific classes from these subsets. Figure \ref{fig:domainnet_visual} offers a glimpse into the synthetic data generated for six domains within the DomainNet dataset. Similar to the ImageNet visualization, each pair of rows represents one of these domains, which encompasses sketch, real, quickdraw, painting, infograph, and clippart domains. Within each column, you will find synthetic images representing individual classes within the respective domain.
Upon close examination, it becomes readily apparent that the synthetic data demonstrates striking color accuracy, precise delineation of object boundaries, and an impressive level of realism that closely approximates that of genuine real-world images. The synthetic images not only maintain vivid and faithful color representations but also capture intricate details, ensuring that the synthetic data closely mirrors the characteristics found in authentic visual data.

\subsubsection{Results on more challenging datasets}
Even for particularly challenging domains such as remote sensing images or fine-grained classification datasets, our method can easily adapt to these scenarios. We conducted experiments on several fine-grained image classification datasets, namely CUB-200~\cite{wah2011caltech}, Stanford Cars~\cite{krause20133d}, and also the satellite image dataset EuroSAT~\cite{helber2019eurosat}. CUB-200 is a challenging dataset consisting of 200 bird species, while Stanford Cars contains 16,185 images belonging to 196 classes of cars. The size of fine-grained recognition datasets is typically smaller compared to general image classification datasets. In previous work~\cite{zhu2022dual,diao2022metaformer}, a common practice is to utilize a pretrained model that has been trained on the ImageNet dataset. In this study, we present two approaches: training the model from scratch and loading a pretrained ResNet34 model.
As shown in Table~\ref{tb:chall}, our method achieves excellent performance even in these challenging domains. Additionally, in the cross-silo federated learning scenario, when clients have strong computational capabilities, one can simply finetune the foundation models on these domains, achieving better performance than normal federated learning methods.



\begin{table*}[]
\centering
\caption{Results for more baselines.}
\label{tb:baselines}
\begin{tabular}{c|ccc|cc}
\toprule
Method                 & FedAvg & FedOpt & Moon  & Ours (one-shot) & Ours (5-round) \\
\midrule
ImageNette  & 72.01  & 73.21  & 74.27 & 85.21           & 93.80          \\
ImageNet100 & 40.13  & 41.25  & 41.43 & 48.31           & 72.67    \\
\bottomrule
\end{tabular}
\end{table*}

\paragraph{Class-level Prompts versus Instance-level Prompts.}  In main experiments, we synthesize a number of samples based on class-level prompts, \eg \emph{A photo of a [class name]}. 
Then following~\cite{lei2023caption}, we caption individual images with foundation models, \eg BLIP-2~\cite{li2023blip}, and synthesize new images based on the individual caption. 
These prompts provide more precise guidance to the generative model by considering the specific characteristics and context of each sample. Note that~\cite{zhou2023training} also produce instance-level prompts by diffusion inversion, while the inverted samples will leak data privacy and it is very expensive to implement diffusion inversion on large datasets for clients.
In Figure~\ref{fig:blip} (Appendix), we present the performances of two kinds of prompts, where we synthesize 1300 images per class, the same number as the real training set. It is evident that employing a more precise instance-level prompt leads to higher accuracy, \eg achieving 78.62\% on ImageNette, compared to class-level prompt, which achieves 73.20\%. This result clearly highlights the importance of considering more detailed instance-level information in prompt design.

\begin{table*}[]
\centering
\caption{Results on challenging datasets.}
\label{tb:chall}
\scalebox{0.7}{
\begin{tabular}{cc|ccc|c|ccc|c}
\toprule
\multicolumn{2}{c|}{Setting} &
  \multicolumn{3}{c|}{FedAvg} &
  \multirow{2}{*}{Ours (one-shot)} &
  \multicolumn{3}{c|}{Ours (5-round)} &
  \multirow{2}{*}{Centralized} \\
  
prompt &
  Dataset &
  $\beta=0.01$ &
  $\beta=0.5$ &
  FedAvg (IID) &
   &
  $\beta=0.01$ &
  $\beta=0.5$ &
  IID &
   \\
   \midrule
class &
  \multirow{2}{*}{EuroSAT} &
  \multirow{2}{*}{43.94} &
  \multirow{2}{*}{74.48} &
  \multirow{2}{*}{84.87} &
  38.37 &
  37.59 &
  82.94 &
  91.01 &
  \multirow{2}{*}{94.3} \\
instance &
   &
   &
   &
   &
  42.7 &
  52.67 &
  92.28 &
  95.26 &
   \\
   \midrule
class &
  \multirow{2}{*}{COVID-19 X-rays} &
  \multirow{2}{*}{57.0} &
  \multirow{2}{*}{86.25} &
  \multirow{2}{*}{95.25} &
  52.0 &
  50.0 &
  70.5 &
  88.75 &
  \multirow{2}{*}{94.5} \\
instance &
   &
   &
   &
   &
  54.75 &
  50.0 &
  64.0 &
  44.0 &
   \\
   \midrule
class & \multirow{2}{*}{BloodMNIST} & \multirow{2}{*}{64.86} & \multirow{2}{*}{83.14} & \multirow{2}{*}{84.2}  & 30.46 & 27.68 & 73.49 & 84.04 & \multirow{2}{*}{84.76} \\
instance &
   &
   &
   &
   &
  30.54 &
  37.1 &
  78.78 &
  83.59 &
   \\
   \midrule
class & \multirow{2}{*}{DermaMNIST} & \multirow{2}{*}{66.65} & \multirow{2}{*}{69.81} & \multirow{2}{*}{71.22} & 34.16 & 45.77 & 64.75 & 73.29 & \multirow{2}{*}{71.26} \\
instance &
   &
   &
   &
   &
  66.79 &
  66.52 &
  69.72 &
  73.59 \\
  \bottomrule
  
\end{tabular}}
\end{table*}

\newpage

\end{document}